\documentclass[letterpaper, 10 pt, conference]{ieeeconf}  

\IEEEoverridecommandlockouts                              
\usepackage{amsmath, nccmath}
\usepackage[numbers]{natbib}
\usepackage{graphicx}
\usepackage{amsfonts}

\usepackage{bbm}

\usepackage{afterpage}
\usepackage{hyperref}
\hypersetup{
    colorlinks=true,
    linkcolor=blue,
    filecolor=magenta,      
    urlcolor=cyan,
    pdfpagemode=FullScreen,
    }

\newcommand{\pp}[1]{\left( #1 \right)}

\newcommand{\mc}{\mathcal}

\title{\LARGE \bf
The boundary of neural network trainability is fractal
}
\author{Jascha Sohl-Dickstein
\\
\tt\small jascha.sohldickstein@gmail.com}

\begin{document}

\maketitle
\thispagestyle{empty}
\pagestyle{empty}

\begin{abstract}
Some fractals -- for instance 
those associated with
the Mandelbrot and quadratic Julia sets -- 
are computed by iterating a function, and identifying the 
boundary between hyperparameters for which the resulting series diverges or remains bounded \citep{mandelbrot1982fractal}. 
Neural network training similarly involves iterating an update function (e.g. repeated steps of gradient descent), can result in convergent or divergent behavior, 
and can be extremely sensitive to small changes in hyperparameters. 
Motivated by these similarities, we experimentally examine the boundary between neural network hyperparameters that lead to stable and divergent training. 
We find that this boundary is fractal 
over more than ten decades of scale in all tested configurations.
\end{abstract}


\section{Introduction}

A common way to generate fractals is to iterate a fixed function $f(\cdot)$ repeatedly, and to keep only the set of points for which small perturbations to hyperparameters\footnote{For consistency with machine learning terminology, I use the term hyperparameter for parameters governing the dynamics or initial conditions of function iteration.} of that function lead to dramatic changes in the sequence of iterated values. 
These points can be thought of as defining a boundary in hyperparameter space along which dynamics bifurcate. For instance, on one side of the boundary function iterations may converge or remain bounded, while on the other side of the boundary they may diverge towards infinity.\footnote{The bifurcation boundary may also be between sequences that converge to different finite solutions (or limit cycles), as in Newton fractals \citep{Tatham_Newton}.}
If the hyperparameters are the initial conditions of the iterated function, this bifurcation boundary is known as the Julia set \citep{julia1918memoire}.

As an example, consider iterating the complex valued function $f(z; c) = z^2 + c$: 
the Mandelbrot fractal \citep{brooks1981dynamics,mandelbrot1982fractal} 
is defined by 
the bifurcation boundary between values of the hyperparameter $c$ for which this iterated function diverges or remains bounded (for an initial $z$ value of 0);
while quadratic Julia sets 
are defined by 
the bifurcation boundary between 
initial $z$ values
for which this iterated functions diverges or remains bounded (for fixed $c$). 



When we train a neural network, we iterate a function (a gradient descent step) of many variables (the parameters of the neural network). For instance, if we perform full batch steepest descent with a fixed learning rate $\eta$, we update the parameters $W$ by iterating the function $f( W; \eta ) = W - \eta\, g( W )$, where $g( W )$ is the gradient of the training loss. 
Iterated steps of gradient descent are known to exhibit bifurcation boundaries, between hyperparameters that lead to converging or diverging training runs. The final loss value achieved when training a neural network has also been shown to have a chaotic dependence on hyperparameters \citep{metz2019understanding,kong2020stochasticity,chen2023stability}.

Motivated by these similarities between fractal generation
and neural network training, in this paper I visualize the bifurcation boundary between hyperparameters which lead to successful and unsuccessful training of neural networks. I find that this boundary is fractal in all experimental conditions, including full batch training with $\operatorname{tanh}$ and $\operatorname{ReLU}$ nonlinearities, training a deep linear network, minibatch training, training on a dataset of size 1, and visualizing training success for different subsets of hyperparameters.


\begin{figure*}[ht]
    \centering
    
    \begin{minipage}{0.47\linewidth}
        \centering
        \textbf{Deep linear full batch (fractal dim 1.17)}
        \href{https://player.vimeo.com/video/903855710?h=0503f33948}{\includegraphics[keepaspectratio,width=\linewidth]{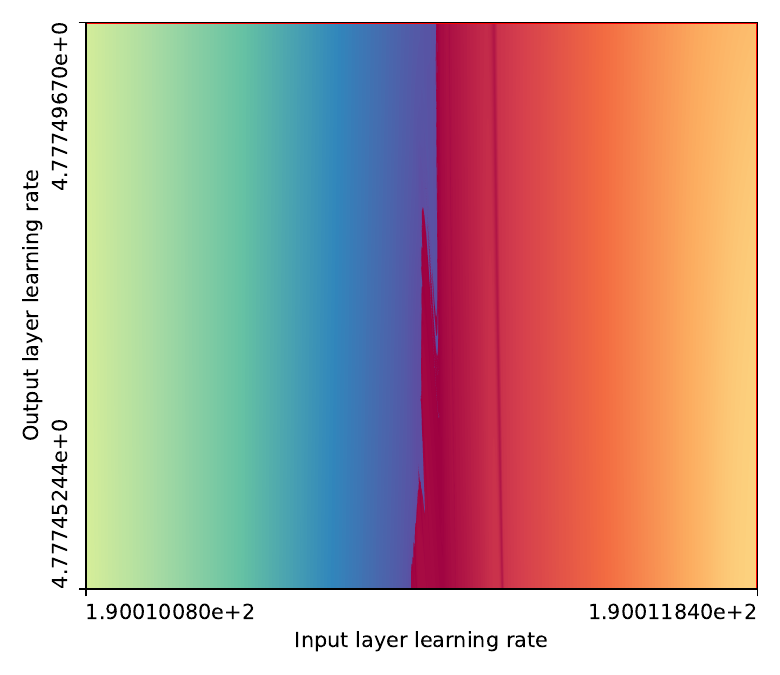}}
    \end{minipage}
    \qquad
    \begin{minipage}{0.47\linewidth}
        \centering
        \textbf{$\operatorname{\mathbf{ReLU}}$ full batch (fractal dim 1.20)}
        \href{https://player.vimeo.com/video/903855690?h=6072a2020e}{\includegraphics[keepaspectratio,width=\linewidth]{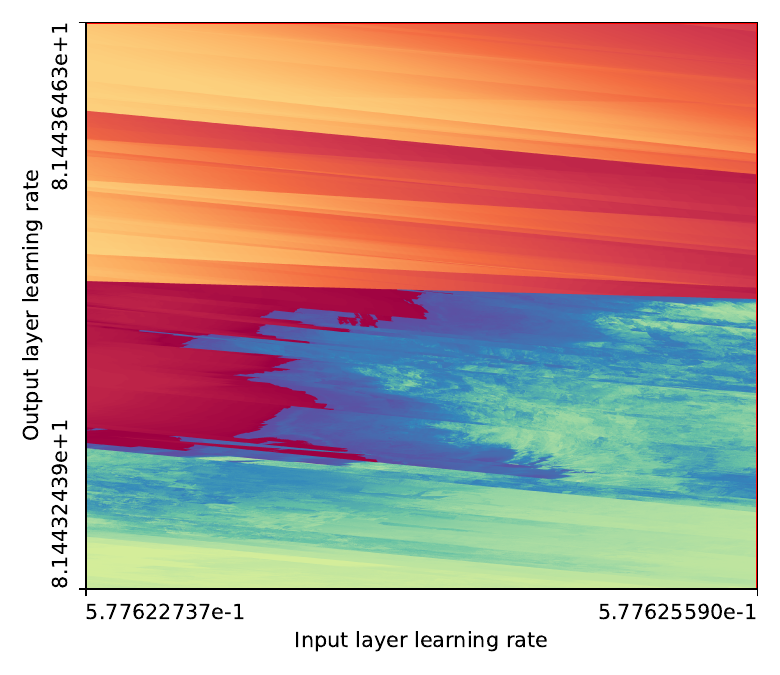}}
    \end{minipage}
    
    \begin{minipage}{0.47\linewidth}
        \centering
        \textbf{$\mathbf{tanh}$ dataset set size 1 (fractal dim 1.41)}
        \href{https://player.vimeo.com/video/904781772?h=8bf3776954}{\includegraphics[keepaspectratio,width=\linewidth]{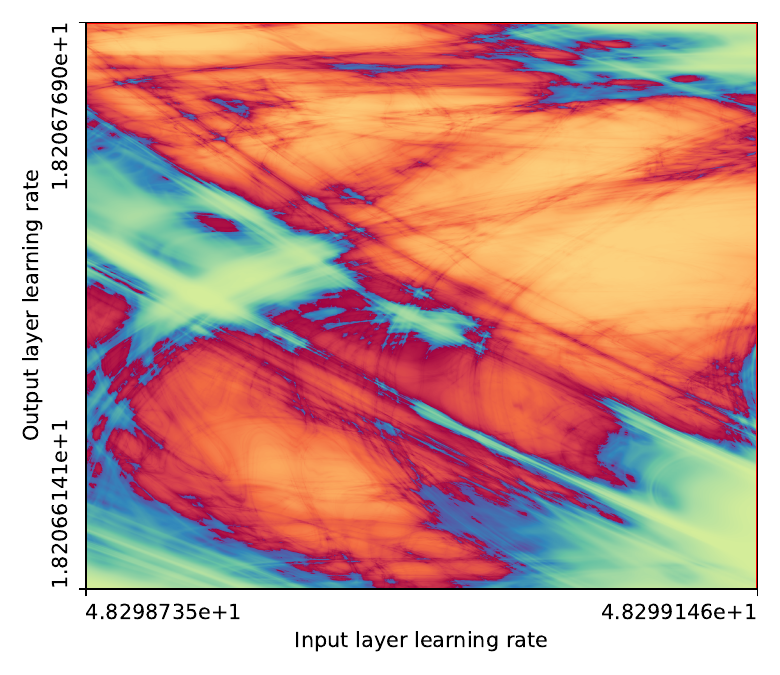}}
    \end{minipage}
    \qquad
    \begin{minipage}{0.47\linewidth}
        \centering
        \textbf{$\operatorname{\mathbf{tanh}}$ minibatch (fractal dim 1.55)}
        \href{https://player.vimeo.com/video/903855680?h=d8f341a934}{\includegraphics[keepaspectratio,width=\linewidth]{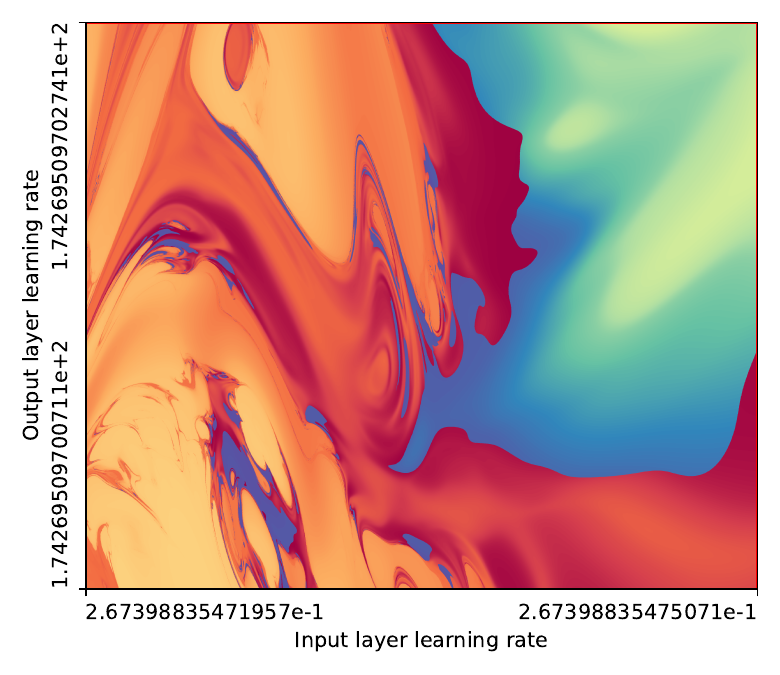}}
    \end{minipage}

    \begin{minipage}{0.47\linewidth}
        \centering
        \textbf{$\operatorname{\mathbf{tanh}}$ full batch (fractal dim 1.66)}
        \href{https://player.vimeo.com/video/903855670?h=ca2b077023}{\includegraphics[keepaspectratio,width=\linewidth]{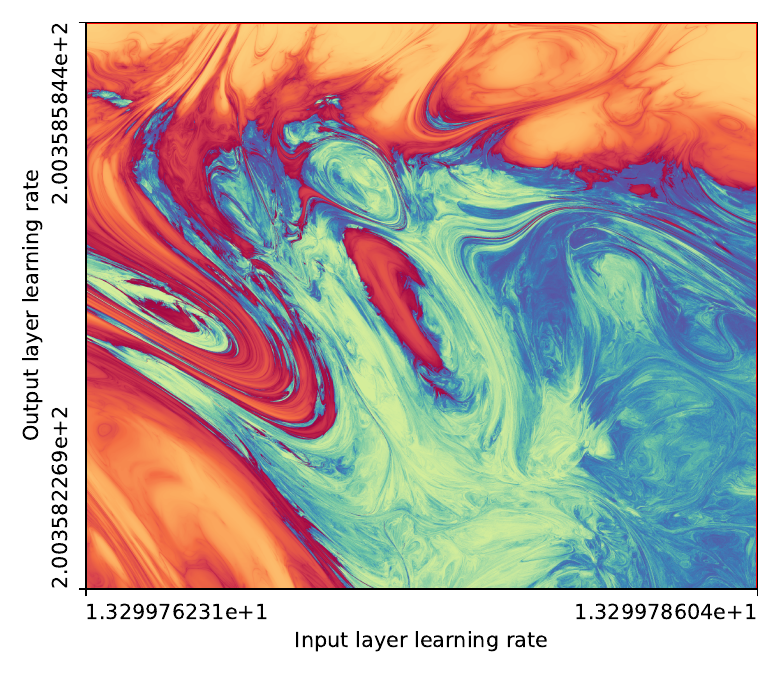}}
    \end{minipage}
    \qquad
    \begin{minipage}{0.47\linewidth}
        \centering
        \textbf{Parameter initialization (fractal dim 1.98)}
        \href{https://player.vimeo.com/video/903855723?h=ed7eb562bb}{\includegraphics[keepaspectratio,width=\linewidth]{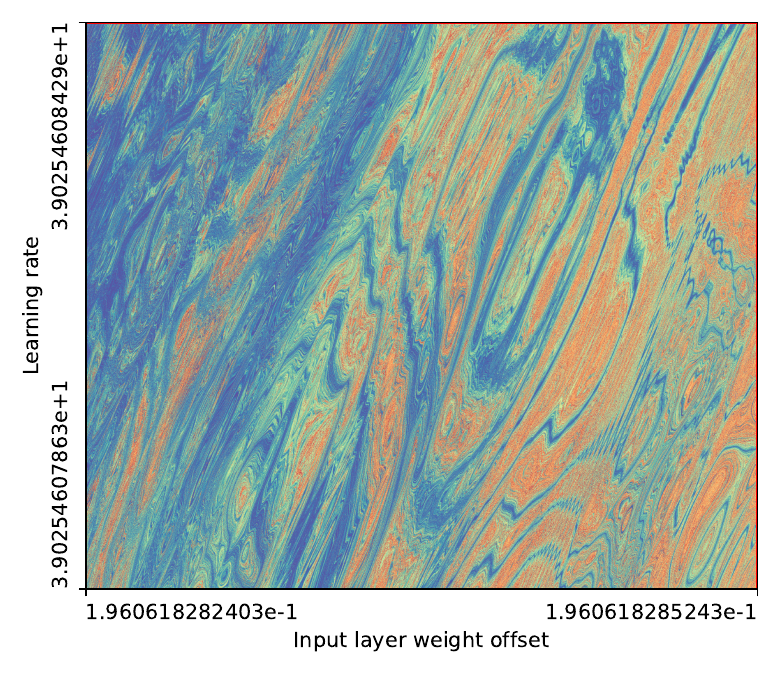}}
    \end{minipage}

    \caption{
    \textbf{The boundary between trainable and untrainable neural network hyperparameters is fractal, for all experimental conditions.} 
    Images show a 2d grid search over neural network hyperparameters.
    For points shaded red, training diverged. For points shaded blue, training converged. 
    Paler points correspond to faster convergence or divergence.
    Experimental conditions include different network nonlinearities, both minibatch and full batch training, and grid searching over either training or initialization hyperparameters. 
    See Section \ref{sec:exp cond} for details. 
    Each image is a hyperlink to an animation zooming into the corresponding fractal landscape (to the depth at which \texttt{float64} discretization artifacts appear). 
    Experimental code, images, and videos are available at
    \url{https://github.com/Sohl-Dickstein/fractal}.   
    }
    \label{fig:experiments}
\end{figure*}
\afterpage{\clearpage}

\section{Experiments}
\label{sec experiments}

\subsection{Network and data}
I train a one hidden layer network with inputs $x \in \mathbb R^n$ and parameters $W_0 \in \mathbb R^{n\times n}$, $W_1 \in \mathbb R^{1\times n}$ on an mse loss,
\begin{align}
    \hat{y}(x; W_0, W_1) &= \alpha_1\, W_1\, \sigma\pp{ \alpha_0 \,W_0 x } \\
    \ell\pp{ W_0, W_1 } &= \frac{1}{|\mathcal{D}|} \sum_{\{x, y\} \in \mathcal{D}} \pp{y - \hat{y}(x; W_0, W_1)}^2
    .
\end{align}
Weights, datapoints, and labels are all randomly initialized from a standard normal distribution, $\mc N\pp{0,1}$. 
The function $\sigma(\cdot): \mc R \rightarrow \mc R$ is applied pointwise. 
The 
scaling factors $\alpha_0$ and $\alpha_1$ are chosen to correspond to the mean field neural network parameterization \citep{mei2018mean}: 
$\alpha_1 = \frac{1}{n}$; 
$\alpha_0 = \sqrt{\frac{2}{n}}$ when $\sigma(\cdot)$ is $\operatorname{tanh}$ or $\operatorname{ReLU}$, and $\alpha_0 = \sqrt{\frac{1}{n}}$ when $\sigma(\cdot)$ is the identity function. The training dataset $\mc D$ has the same number of examples $\left|\mc D\right|$ as the number of free parameters of $f(\cdot)$. $\left|\mc D\right| = \pp{n^2 + n}$ datapoints for nonlinear networks, and $\left|\mc D\right| = n$ datapoints for deep linear networks. Input and hidden layer widths are fixed to $n=16$.

\subsection{Training}


The input and output weights $W_0$ and $W_1$ are trained with learning rates $\eta_0$ and $\eta_1$ respectively. 
Training consists of 500 (sometimes 1000) iterations of full batch steepest gradient descent. 
Training is performed for a 2d grid of $\eta_0$ and $\eta_1$ hyperparameter values, with all other hyperparameters held fixed (including network initialization and training data). Training was performed in \texttt{float64}.
Some of these design choices are modified for individual experiments, as stated.


\subsection{Visualization and analysis}
\label{sec vis}
Training runs that diverge are shown in shades of red. Training runs that converge are shown in shades of blue. For converging runs, color intensity is proportional to $\sum_t \ell_t\pp{\cdot}$, where $\ell_t$ is the loss at training step $t$. The more intense the blue color, the longer the training run spent with higher loss. For diverging runs, color intensity is proportional to $\sum_t \ell_t^{-1}$. The more intense the red, the longer the training run spent with lower loss. The color scale is adapted to each image.

For each zoom sequence, a series of roughly 50 images of size $4096\times 4096$ pixels are generated, each of which increases the zoom factor by a factor of two. The zoom animation interpolates between these images.

The estimated fractal dimension in Figure \ref{fig:experiments} is the median of the estimated fractal dimension for all $\sim$50 images in the zoom sequence. Estimation is performed using the boxcount method of PoreSpy \citep{gostick2019porespy}. 

\subsection{Experimental conditions}
\label{sec:exp cond}

To explore how general fractal behavior is, I performed experiments in six conditions:
    \begin{enumerate}
        \item Baseline condition: $\operatorname{tanh}$ nonlinearity, full batch gradient descent, grid search over $\eta_0$ and $\eta_1$ learning rate hyperparameters.
        \item $\operatorname{ReLU}$ nonlinearity.
        \item Identity nonlinearity (i.e., a deep linear network).
        \item Minibatch gradient descent, with minibatch size 16.
        \item Only a single training datapoint, $\left| \mathcal{D} \right| = 1$.
        \item A grid search over a different pair of hyperparameters, one of which specified the mean value used during parameter initialization, and the other of which specified the learning rate used for both parameters.
    \end{enumerate}
Unspecified design choices are the same as in the baseline condition, and as described earlier in Section \ref{sec experiments}. Representative images from all experimental conditions are shown in Figure~\ref{fig:experiments}, sorted by fractal dimension, and with links to the corresponding animations.


\section{Discussion}

\subsection{Elaborate functions in high dimensional spaces}

Most popular fractals defined by bifurcation boundaries 
iterate only a 
simple one-dimensional function, consisting of a low degree polynomial or ratio of polynomials \citep{brooks1981dynamics,mandelbrot1982fractal,michelitsch1992burning,markus1998lyapunov,milnor2011dynamics,Tatham_Newton}. 
The resulting fractals are typically perceived as possessing a lot of both repeated geometric structure and symmetry (e.g. consider the presence of `mini-Mandelbrot' sets deep within the Mandelbrot set).

In contrast, neural network training involves iterating a complicated function, with many random terms stemming from weight initialization and training data, acting in a high dimensional space (i.e. the function acts on the parameter space of the neural network being trained).
The resulting fractals seem visually more organic, with less repeated structure and symmetry. 
It will be a fascinating to further explore how properties of fractals depend on properties of the generating function.

\subsection{Non-homogeneity of boundary}

It will similarly be fascinating to explore how properties of the bifurcation boundary vary for a single generating function, in different regions of hyperparameter space. 
I decided which regions of the hyperparameter landscape to explore by hand in an ad hoc way, and the resulting images are inevitably biased. 
A limiting example to consider is when the learning rate $\eta_0$ for the input layer is made very small, so that only the readout layer trains. 
Training only the readout layer corresponds to linear regression on an mse loss, with dynamics that are known in closed form \citep{boyd2004convex}, and are not fractal. 
So some regions of the bifurcation boundary for the experimental conditions in Section \ref{sec experiments} will not be fractal.


\subsection{Stochastic training}

For minibatch training, the iterated function is stochastic rather than deterministic due to minibatch sampling. 
I was surprised that this stochastic function also generated fractals, without the fine multiscale structure being corrupted by minibatch noise. 
This is suggestive of Lyapunov fractals \citep{markus1998lyapunov}, for which the function being iterated changes at every time step in a sequence, though in a more restricted way. 

\subsection{Higher dimensional fractals}

This paper explored fractals in the two dimensional space defined by pairs of hyperparameters. Neural network training involves countless hyperparameters (e.g. we could specify an initialization, learning rate schedule, and regularization schedule for every weight in the network, in addition to data augmentation and loss function hyperparameters). 
It has been a challenge to extend Mandelbrot and Julia sets to higher dimensions in a satisfying way. 
That challenge should not exist for fractals stemming from neural network hyperparameters; they are naturally defined in three or more dimensions. 

\subsection{Meta-loss landscapes are difficult to navigate}

Many types of meta-learning optimize hyperparameters associated with neural network training (e.g. this is done in learned optimizers \citep{metz2022velo}). 
The meta-loss landscapes associated with neural network hyperparameters are often pathological and chaotic, and descending this badly behaved landscape is a central challenge in meta-learning \citep{metz2019understanding}.

The loss functions visualized in Figure \ref{fig:experiments} can be interpreted as meta-loss landscapes. These experiments therefore suggest a more nuanced explanation for chaotic meta-loss landscapes; meta-loss landscapes have extreme sensitivity to small changes in hyperparameters, because they are fractal in those hyperparameters.

Although we only observe fractal structure at the boundary between hyperparameters that result in successful or failed neural network training, this is nonetheless relevant for meta-learning. The best performing hyperparameters are typically near the edge of instability, and meta-training seeks out this region in order to minimize the meta-loss.

\subsection{Fractals are beautiful and relaxing}

This has been a particularly fun project to work on (and the first project where my daughter is as excited about the results as I am). 
I hope the reader found these experiments unusually enjoyable, as I did.

\small

\section*{Acknowledgments}

Thank you to Maika Mars Miyakawa Sohl-Dickstein for inspiring the original idea, and for detailed feedback on the 
generated fractals.

\bibliographystyle{plainnat}
\bibliography{fractal_neural_nets.bib}


\end{document}